%% file: iclp2014_final.tex
\documentclass[supp]{new_tlp}

\usepackage{supp}%%added

\usepackage{eurosym}
\usepackage{graphicx}
\usepackage{subcaption}
\usepackage[smaller,nolist]{acronym}
\usepackage{enumerate}
\usepackage{hyperref}
\usepackage{amsmath}

\usepackage{globalopt}

  \title[Multi-Criteria Optimal Planning for Energy Policies in CLP] 
        {Multi-Criteria Optimal Planning for Energy Policies in CLP}

  \author[M. Gavanelli et al.]
         {MARCO GAVANELLI\\
         EnDiF - Universit\`a di Ferrara, Italy\\
         \email{marco.gavanelli@unife.it}
		\and STEFANO BRAGAGLIA\\
			Department of Computer Science, University of Bristol, UK\\
			\email{stefano.bragaglia@bristol.ac.uk}
		\and MICHELA MILANO, FEDERICO CHESANI\\
			DISI - Universit\`a di Bologna, Italy\\
			\email{\{michela.milano | federico.chesani\}@unibo.it}
		\and ELISA MARENGO\\
			Faculty of Computer Science - Free University of Bozen-Bolzano\\
			\email{elisa.marengo@unibz.it}
		\and PAOLO CAGNOLI\\
         ARPA Emilia-Romagna, Italy \\
         	\email{PCagnoli@arpa.emr.it}
         }

\jdate{March 2003}
\submitted{14 February 2014}
\revised{18 April 2014}
\accepted{15 May 2014}
\pubyear{2003}
\pagerange{\pageref{firstpage}--\pageref{lastpage}}
\doi{S1471068401001193}
\pubauthor{M. Gavanelli et al.}%%added
\jurl{xxxxxx}
\pubdate{22 June 2013}

\begin{document}

\label{firstpage}

\maketitle

\begin{abstract}
In the policy making process a number of disparate and diverse issues such as economic development, environmental aspects, as well as the social acceptance of the policy, need to be considered. A single person might not have all the required expertises, and decision support systems featuring optimization components can help to assess policies.

Leveraging on previous work on Strategic Environmental Assessment, we developed a fully-fledged system that is able to provide optimal plans with respect to a given objective, to perform multi-objective optimization and provide sets of Pareto optimal plans, and to visually compare them.
Each plan is environmentally assessed and its footprint is evaluated.
The heart of the system is an application developed in a popular Constraint Logic Programming system on the Reals sort.
It has been equipped with a web service module that can be queried through standard interfaces, and an intuitive graphic user interface.

	%%%The number of issues, stakeholders needs and regulations that a policy must consider in the current world
	%%%is so high that addressing them only with common-sense is unthinkable.
	%%%Policy makers have to consider disparate issues, as diverse as economic development,
	%%%environmental aspects, as well as the social acceptance of the policy.
	%%%A single person cannot be expert in all these subjects. Thus, to obtain a well assessed policy in the current complex world one can adopt decision support systems featuring optimization components.
	%%%
	%%%Leveraging on previous work on Strategic Environmental Assessment, 
	%%%we developed a fully-fledged
	%%%system that is able to provide optimal plans with respect to a given objective, 
	%%%to perform multi-objective optimization and provide
	%%%sets of Pareto optimal plans,
	%%%and to visually compare them.
	%%%Each plan is environmentally assessed and its environmental footprint is provided in terms of emissions,
	%%%global warming effect, human toxicity, and acidification.
	%%%The heart of the system is an application developed in a popular Constraint Logic Programming 
	%%%system on the Reals sort.
%%%It has been equipped with a web service module that can be queried
%%%through standard interfaces. An intuitive graphic user
%%%interface has been added to provide easy access to the web service and
%%%the CLP application.
  \end{abstract}


%\tableofcontents

\input{intro}

\input{model}

\input{emissions}

\input{aggregators}

\input{pareto}

\input{gui}

%\input{related}

\input{conclusions}

\paragraph{\textbf{Acknowledgements.}} This work was partially supported by EU project ePolicy, FP7-ICT-2011-7, grant agreement 288147.

%%%\section*{Acknowledgements}
%%%This work was partially supported by EU project ePolicy, FP7-ICT-2011-7, 
%%%grant agreement 288147.
%%%Possible inaccuracies of information are under the
%%%responsibility of the project team. The text reflects solely the views of its
%%%authors. The European Commission is not liable for any use that may be
%%%made of the information contained in this paper.

%%%\appendix
%%%
%%%\input{paretoAppendix}
%%%
%%%\input{guiAppendix}

\bibliographystyle{acmtrans}
\bibliography{bib,biomasse,odessa}

\label{lastpage}
\input{acronyms}
\end{document}

%% file: intro.tex
\section{Introduction}
\label{sec:introduction}

Policy making, in the current connected world, has to consider such a number of issues that 
a single person cannot possibly consider without introducing vast approximations.
For example European regions should provide Regional Energy Plans to define strategic objectives and political actions for the energy sector, considering:
\begin{itemize}
\item the current energy balance in the region (produced/consumed energy, imported/ex-ported, electrical/thermal, etc.)
\item forecasts for the following years, about energy request or production costs;
\item existing and new directives, e.g. the EU 20-20-20 initiative that poses three challenging 
targets for 2020:
20\% improvement of energy efficiency, 20\% of the energy produced from renewable sources, and 20\% reduction of greenhouse gas emissions.
\end{itemize}

The policy contains strategic objectives on the energy share and energy efficiency, measures and activities to cope with the increased energy needs, new regulations, etc. Regional plans in particular are typically very high-level: they include activities such as building new power plants for some total output power, the share of each fuel type (nuclear, fossil fuels, biomasses, etc.) and the type of produced energy (electric or thermal); but they lack information about, for example, the actual placement of the plants in the region, since more detailed plans will be done at lower scale, like the province or municipality levels. By EU directives, regional policies on the energy sector should also include an environmental assessment of the plan. Being the plan so high-level, usually the assessment is done only in a qualitative way.

In a previous work \cite{GavRigMilCag10-ICLP10-IJ}, we proposed and compared two alternative logic programming formulations for the strategic environmental assessment of regional plans; one was based on probabilistic logic programming, the other on \ac{CLP} \cite{JM94}. We also developed four fuzzy-logic formulations of the assessment problem \cite{NostroAIIA}. All these programs consider a regional plan, given in input, and provide its environmental assessment. 
In a following work \cite{GavRigMilCag12-CIDASD12-BC}, the \ac{CLP} program was extended to generate plans together with their assessment, and it was used during the definition of the Regional Energy Plan 2011-2013 of the Emilia-Romagna region \cite{PER2011-13}.

In this work, we show how the first prototype of the planner was extended to a fully-fledged application.
In particular, the current version of the software supports:
\begin{itemize}
	\item plans that consider decommissioning obsolete power plants;
	\item computation of emissions of the power plants for various types of pollutants, in a quantitative way;
	\item quantitative assessment of the effect of the plan on human health, global warming, and acidification potential;
	\item multi-criteria optimization considering a variety of objective functions based on qualitative and quantitative information;
	\item computation of the Pareto front, for two or more objective functions;
	\item a web service, providing access through a \ac{GUI} and APIs.
\end{itemize}

This work is one of the components of the EU ePolicy project\footnote{\url{http://www.epolicy-project.eu}}.
The final application will include also an opinion mining component, to assess the acceptance of the policies from the public considering information coming from blogs and social networks; a social simulator component, that will simulate how the population will react to the policies adopted by the Region; a mechanism design component, that will include information from game theory to provide the best allocation schemes of regional subsidies to the stakeholders; and an integrated visualization component.

The rest of the paper is organized as follows.
We first introduce the planning and environmental assessment as they are currently done by experts in the Emilia-Romagna region of Italy, and recap the basic \ac{CLP} program of the first prototype (Section~\ref{sec:model}).
In Section~\ref{sec:extensions}, we extend it with new features.
We show the design and features of the web service and \ac{GUI} in Section~\ref{sec:gui}.
Finally, we conclude in Section~\ref{sec:conclusions}.

%%% Local Variables:
%%% TeX-master: "iclp14"
%%% End:

%% file: model.tex
\section{Problem considered and CLP solution}
\label{sec:model}

The strategic environmental assessment, in the Emilia-Romagna region of Italy,
is currently performed by considering two matrices, called {\em coaxial matrices} \cite{LibroCagnoli}.
They are a development of the network method \cite{Sorensen}, and they contain qualitative relations.

The first matrix, \Mop, considers the {\em activities} that can be undertaken in a plan, and links them with the {\em environmental pressures}. % or {\em impacts}.
Pressures can be positive or negative, and they account for the impact on the environment of human activities.
Each element $\mop^i_j$ of the
matrix \Mop\ can take values $\{${\em high}, {\em medium}, {\em low}, {\em null}$\}$,
and defines a qualitative dependency between the
activity $i$ and the negative or positive pressure $j$.

The second matrix, \Mpr, relates the pressures with the {\em environmental receptors}, that register the effect of the pressures on the environment. For example, the activity {\em ``coal-fueled power plant''} generates the pressure {\em ``emission of pollutants in the atmosphere''}; on its turn, this influences the receptor {\em ``air quality''} (as well as other receptors, like e.g. {\em ``human wellbeing''}). Each element $\mpr^i_j$ of the matrix can take the qualitative values: {\em high}, {\em medium}, {\em low} or {\em null}, and defines the dependency between pressure $i$ and receptor $j$.

Currently, the matrices relate 115 activities with 29 negative and 19 positive pressures, and 23 receptors. They can be used to assess a variety of regional plans, including Agriculture, Forest, Fishing, Energy, Industrial, Transport, Waste, Water, Telecommunications, Tourism, Urban plans.
The environmental assessment is usually done using a spreadsheet and deleting (by hand) those activities that do not belong to the given type of plan; then pressures and receptors that are not influenced by remaining activities are removed accordingly. The ``reduced'' matrices are evaluated by environmental experts, that state which parts are most important, mainly considering clusters of {\em High} values.

Clearly, this process is very slow, experts might overlook important combinations of {\em medium} or {\em low} values, and, most importantly, it can be done only after the plan has been provided by the policy maker. 
At this stage, usually only minor modifications can be back-propagated to the plan, and comparing a plan's effects with alternative plans is not possible without starting another planning phase.

\subsection{A CLP solution}
\label{sec:model_recap}

To overcome the limitations and improve on current practices, we devised a \ac{DSS} able to provide optimal plans and environmental assessment \cite{GavRigMilCag12-CIDASD12-BC}: the planning problem was modelled as a linear program in \ac{CLP} on the Reals sort (\clpr).

Given a number \Nope\ of
activities, we consider a vector $\Ope =(\ope_1,\dots, \ope_\Nope)$ in which we associate to each activity a variable $\ope_i$ that defines its magnitude. 
The domain of $\ope_i$ depends on the availability of the resource on the given Region; for example some regions are very windy, while others can exploit better biomasses or solar energy.

We distinguish primary from secondary activities: primary activities are directly related to the given type of plan, while secondary ones are those supporting the primary activities by providing the needed infrastructures. E.g. in an energy plan, primary activities are those producing energy (e.g., power plants),
and they may require other activities (e.g., power lines, waste stocking, streets, etc.) that have an environmental impact too.
Let $\PrimaryActivities$ be the set of indexes of primary activities and $\SecondaryActivities$ that of secondary activities. The dependencies between primary and secondary activities are considered by the constraint:
\begin{equation}
 \forall j \in \SecondaryActivities  \ \ \ \ope_j = \sum_{i\in \PrimaryActivities} \dep_{ij} \ope_i 
 \label{eq:primary_secondary}
\end{equation}
Each activity $\ope_i$ has a cost $\cost_i$; given a budget $\budget_{Plan}$ available for a given plan, we have:
\begin{equation}
 \sum_{i=1}^\Nope \ope_i \  \cost_i \leq \budget_{Plan}
 \label{eq:budget}
\end{equation}
Given an expected outcome $\out_{Plan}$ of the plan, we also have:
\begin{equation}
 \sum_{i=1}^\Nope \ope_i \   \out_i \geq \out_{Plan}.
\end{equation}
E.g. an energy plan outcome can be to increase available
energy, so $\out_{Plan}$ could be the added availability of electrical
power (in kilo-TOE, Tonnes of Oil Equivalent). Other outcomes
can be considered, e.g. increasing only renewable energies.

Concerning the impacts of the regional plan, an environmental expert suggested to convert the qualitative values in the matrices into coefficients ranging from 0 to 1; we sum up the contributions of all the activities to estimate the impact on each pressure:
\begin{equation}
 \forall j \in \{1,\dots,\Npre\} \quad  \pre_j =\sum_{i=1}^\Nope \mop^i_j  \  \ope_i.
\label{eq:pre_ope}
\end{equation}
Similarly, given the pressures $\Pre=(\pre_1,\dots,\pre_\Npre)$, the influence on the environmental receptor $\ric_i$ is estimated through the matrix \Mpr, relating pressures with receptors:
\begin{equation}
 \forall j \in \{1,\dots,\Nric\} \quad \ric_j = \sum_{i=1}^\Npre \mpr^i_j \pre_i.
\label{eq:ric_pre}
\end{equation}

Possible objective functions include maximizing/minimizing the produced energy,
the cost, or one of the receptors (e.g., {\em ``air quality''}),
or a linear combination of the above.

\section{Extended solution}
\label{sec:extensions}

The \clpr\ program described in Section~\ref{sec:model_recap} was used in the development of the 2011-13 Regional Energy plan of the Emilia-Romagna region of Italy: the plan objective was to increase the share of renewable energy in the energy mix, and to fulfil the 20-20-20 directive.
For the next years experts foresee the decommissioning of carbon-based power plants, with a residual utilisation when renewable energy is unavailable or in peak hours. Region experts asked us to extend the \ac{DSS} to consider also the closing of power plants.

Power plants decomissioning implies that some activities have a
negative magnitude: e.g., the magnitude, in MW, of oil-based power
plants could be reduced w.r.t. the previous years. However, negative
activities introduce non-linearities. For example, if building a new
plant $i$ has a cost $\cost_i$ in \euro$/MW$, decommisioning  it will not give a {\em profit} of $\cost_i$ \euro$/MW$.

Our implementation is based on the \eclipse\ CLP language \cite{AptWallace_ECLiPSe_book,ECLiPSe_TPLP}, using the {\tt eplex} library \cite{eplex}. The {\tt eplex} library uses very fast solvers using linear programming or mixed-integer linear programming algorithms, allowing the use of linear constraints on variables ranging either on continuous or on integer domains.
It is well known that linear programming is polynomially solvable, while
(mixed) integer linear programming is NP-hard; thus
the efficiency of the solution depends on whether there are integer variables or not.
To address the non-linearity, we introduced, for each
activity $\ope_i$ that has negative values in its domain,
a real variable $\PosActivity_i$ defined as:
$$
\PosActivity_i = \left\{
	\begin{array}{rcl}
		\ope_i 	&	\mbox{if}	& \ope_i \geq 0 \\
		0	&	\mbox{if}	& \ope_i < 0
	\end{array}
	\right.	
$$
The cost constraint (\ref{eq:budget}) is now rewritten as:
\begin{equation}
 \sum_{i=1}^\Nope \PosActivity_i \  \cost_i \leq \budget_{Plan}.
 \label{eq:cost_constraint_integer}
\end{equation}
Similarly, secondary activities should not be decommisisoned together with primary ones; 
so we impose their relationship only with the positive part of primary activities.

Concerning the environmental assessment, we may notice that any new activity has different types of impacts, some related to its initial implementation (e.g., land use for building a coal power plant), and others due to the activity functioning (e.g., air pollution for burning fuel). Equation (\ref{eq:pre_ope}) correctly accounts for both when dealing with ``positive'' activities, while would be incorrect w.r.t. ``negative'' activities.

To cope with the activities decommissioning, the co-axial matrices have been extended with new activities (e.g., {\em ``Reduced use of fossil fuelled power plants''}). All the pressures are now computed  on positive activities only, and Equation~(\ref{eq:pre_ope}) is substituted with
\begin{equation}
 \forall j \in \{1,\dots,\Npre\} \quad  \pre_j =\sum_{i=1}^\Nope \mop^i_j  \  \PosActivity_i.
\label{eq:pre_ope_new}
\end{equation}
We considered the new activities as a new type of secondary activities: the {\em ``Reduced use of fossil fuelled power plants''} is a secondary activity that becomes positive only when activities like
{\em ``Coal-based power plant''}, etc., has a negative value (i.e., in case of decommissions).
Hence, we now have two matrices of dependencies between activities: a $\Nope \times \Nope$ square matrix $\Dep^+$  where each element $\dep_{ij}^+$ represents the magnitude of activity $j$ per unit of activity $i$;
and another $\Nope \times \Nope$ square matrix $\Dep^-$ where each element $\dep_{ij}^-$ represents the magnitude of activity $j$ per unit of reduction of activity $i$. Equation~(\ref{eq:primary_secondary}) is updated with
\begin{equation}
\begin{array}{rcl}
 \forall j \in \SecondaryActivities  \ \ \ \ope_j = \sum_{i\in \PrimaryActivities} K_{ij}
 \label{eq:primary_secondary_new}
 &
 ~
 &
K_{ij} = \left\{
	\begin{array}{lll}
		 \dep_{ij}^+ \cdot \ope_i & \mbox{if} & \ope_i \geq 0 \\	
		 \dep_{ij}^- \cdot (-\ope_i) & \mbox{if} & \ope_i < 0 \\
	\end{array}
	\right. 
\end{array}
\end{equation}

%%% Local Variables:
%%% TeX-master: "iclp14"
%%% End:

%% file: emissions.tex
\subsection{Computing emissions}
\label{sec:emissions}

A further extension to the model presented in Section~\ref{sec:model_recap} has been about the evaluation of emissions in quantitative terms. To this end, we rely on the data provided by two databases: INEMAR \cite{INEMAR} and ISPRA \citeA{ISPRA}: both databases provide the various types of pollutants
\footnote{Considered types of pollutants include \ac{SOx}, \ac{NOx},
 methane, $CO$, $CO_2$, $N_2O$, ammonia, \ac{HCB},
various metals (Arsenic,	Cadmium,	Chromium, Copper, Mercury,	Nickel,	lead,	Selenium,	Zinc),
particulate matter (PM10), Dioxins, and some families of compounds, like \ac{PAH}, \ac{PCB}, and \ac{NMVOC}.}
emitted per fuel unit (in GJ). While ISPRA provides the average emission for each plant type, INEMAR provides fine grained information, in which also the type of boiler and the size of the plant (in MW) are considered.

Let \Ncald\ the number of boiler types, and $\Cald =(\cald_1,\dots, \cald_\Ncald)$ a vector of constrained variables where $\cald_i$ is the total output power of plants using boiler type $i$. Let $\Moc$ be the matrix that relates power plants and the different kinds of boiler: each element $\moc^i_j$ of the matrix is set to  if the boiler $\cald_j \in \Cald$ can be used for the power plant $\ope_i \in \Ope$, and zero otherwise.
The output power of each plant type is the sum of the power of its boilers:
\begin{equation}
 \forall i \in \{1,\dots,\Nope\}   \ \ \ \ope_i = \sum_{j\in \Ncald} \moc^i_j \cald_j
 \label{eq:activity_caldaie}
\end{equation}

Let $\Emi =(\emi_1,\dots, \emi_\Nemi)$ be the vector of emissions and $\Mec$ the matrix relating them with 
the boilers. An element $\mec^i_j \in \Mec$ represents the grams of pollutant $\emi_i \in \Emi$ 
emitted when 1GJ of fuel is provided to the boiler $\cald_j \in \Cald$. 
To calculate the emissions, we have to compute the input energy for each boiler type $j$, provided the output power
$\cald_j$:
\begin{equation}
 \forall i \in \{1,\dots,\Nemi\}   \ \ \ \emi_i = \sum_{j\in \Ncald} \mec^i_j \left( \frac{\hoursPerYear}{\efficiency} \cald_j\right).
 \label{eq:emission_caldaie}
\end{equation}
\hoursPerYear\ is the average running time of a power plant per year (necessary to convert energy into power) and \efficiency\ is the average efficiency (output power/input power) of power plants, which is prescribed by law as 39\% \cite{EfficienzaMediaCentrali}.

%%% Local Variables:
%%% TeX-master: "iclp14"
%%% End:

%% file: aggregators.tex
\subsection{Indicators}
\label{sec:aggregators}

Thanks to the extension presented in Section~\ref{sec:emissions} it is possible to evaluate quantitatively the emissions of some gases, metals, etc. Such data can be exploited to consider plans aiming to minimize them; however, it is not clear how to compare the emissions. E.g., a policy maker could know that \ac{NOx} are toxic for humans, but how does that compare with heavy metal emissions?

The \citeN{BREF} published a set of indicators quantifying the effect of various substances on {\em human toxicity}, {\em global warming} and {\em acidification}: e.g., the Annex 1 contains 100 chemicals with their human toxicity factor, defined as the toxicity of the substance compared to that of lead (Pb).
By using the weights in the tables, one can provide, e.g., the total human toxicity (in kg of {\em equivalent emitted Pb}), the global warming effect (kg of {\em equiv. $CO_2$}) and the acidification of the plan (kg of {\em equiv. $SO_2$}). Moreover, a policy maker may want to optimize on these indicators (by directly minimizing them or any weighted sum).

However, the tables provided by the EC do not always have the same granularity of the information available for emissions. For example, for each plant type we know the emissions of \ac{NOx}, while in the EU report there are the single toxicity values of $NO$ and $NO_2$ (and they are quite different: respectively, 95 and 300 times that of Pb).
Environmental experts suggested to provided as output, for each indicator, the best, worst, and average cases, considering respectively the highest toxicity in the compound class, the lowest and an average.
If one of the indicators is in the objective function (e.g., one wants to find the plan with minimum human toxicity), we optimize the worst case.

%%% Local Variables:
%%% TeX-master: "iclp14"
%%% End:

%% file: pareto.tex
\newcommand{\mobj}{\mu}
\newcommand{\utopia}{\mobj^u}
\newcommand{\nadir}{\mobj^N}
\newcommand{\normal}{\overline{\mobj}}

\subsection{Computing the Pareto front}
\label{sec:pareto}

The approach presented in Section~\ref{sec:model} allows to optimize w.r.t. a single function. However, in the case of regional planning it is very hard (if not impossible) to devise a unique function
that includes all the objectives that are important for the user. Hence, we added a further extension towards multi-objective optimization.

In a multi-objective optimization problem, a solution is {\em Pareto optimal} if it is not possible to improve the result for one objective function, without worsening at least another objective function.
More precisely, in a multi-objective problem with $n$ functions to minimize, a solution $\mobj^*$ is \emph{Pareto-optimal} if there does not exist another solution $\mobj$ such that $\mobj_j \leq \mobj_j^*$ for $1 \leq j \leq n$ and there exists at least one $i$, $1 \leq i \leq n$ such that $\mobj_i < \mobj_i^*$.
The set of Pareto points is distributed on the so-called Pareto frontier.

We implemented the {\em normalized normal constraint method} \cite{nnc}, an algorithm that works with any type of constraints (linear and nonlinear) and variables (continuous and discrete), and that is able to find an \emph{evenly distributed} set of Pareto solutions. In this way, the policy maker is provided with a set of solutions that are a good representation of the whole space of the Pareto frontier.

%%% Local Variables:
%%% TeX-master: "iclp14"
%%% End:

%% file: gui.tex
\section{Graphical User Interface}
\label{sec:gui}

\begin{figure}[tb]
        \centering
        \begin{subfigure}[b]{0.34\textwidth}
                \includegraphics[width=\textwidth]{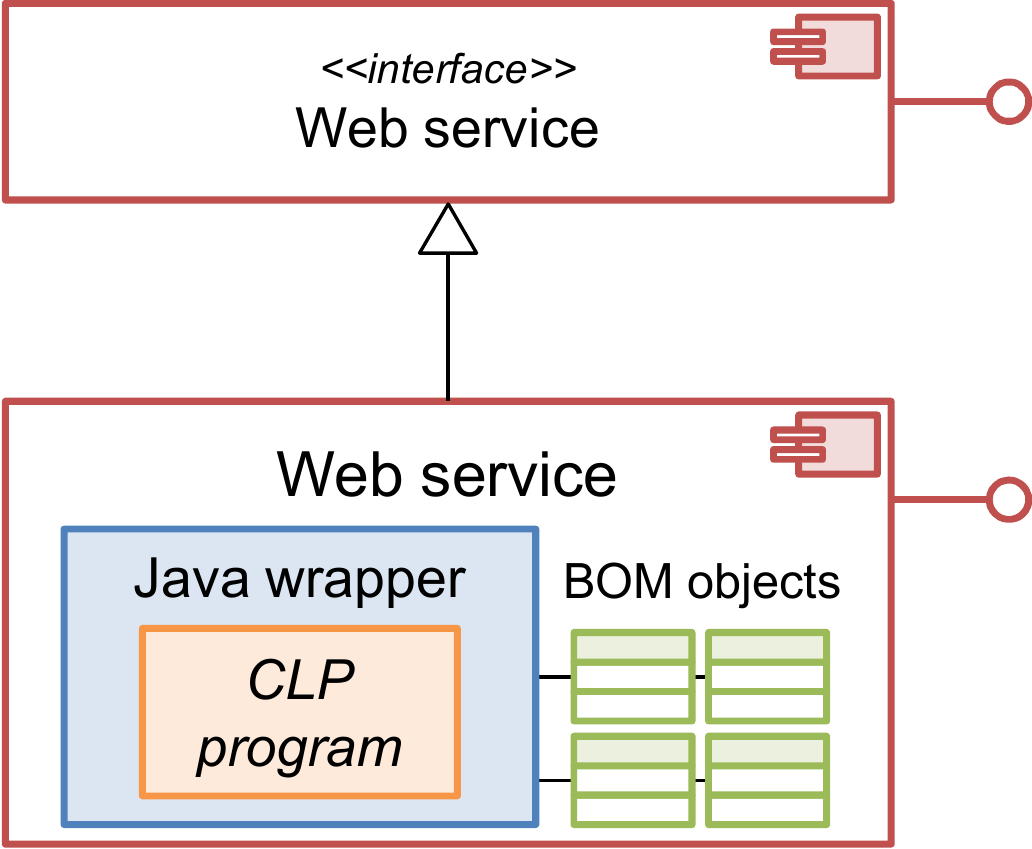}
                \caption{The Web service.}
				\label{fig:service}
        \end{subfigure}%
        ~~~~~~~~~~ %add desired spacing between images, e. g. ~, \quad, \qquad etc.
        \begin{subfigure}[b]{0.5\textwidth}
                \includegraphics[width=\textwidth]{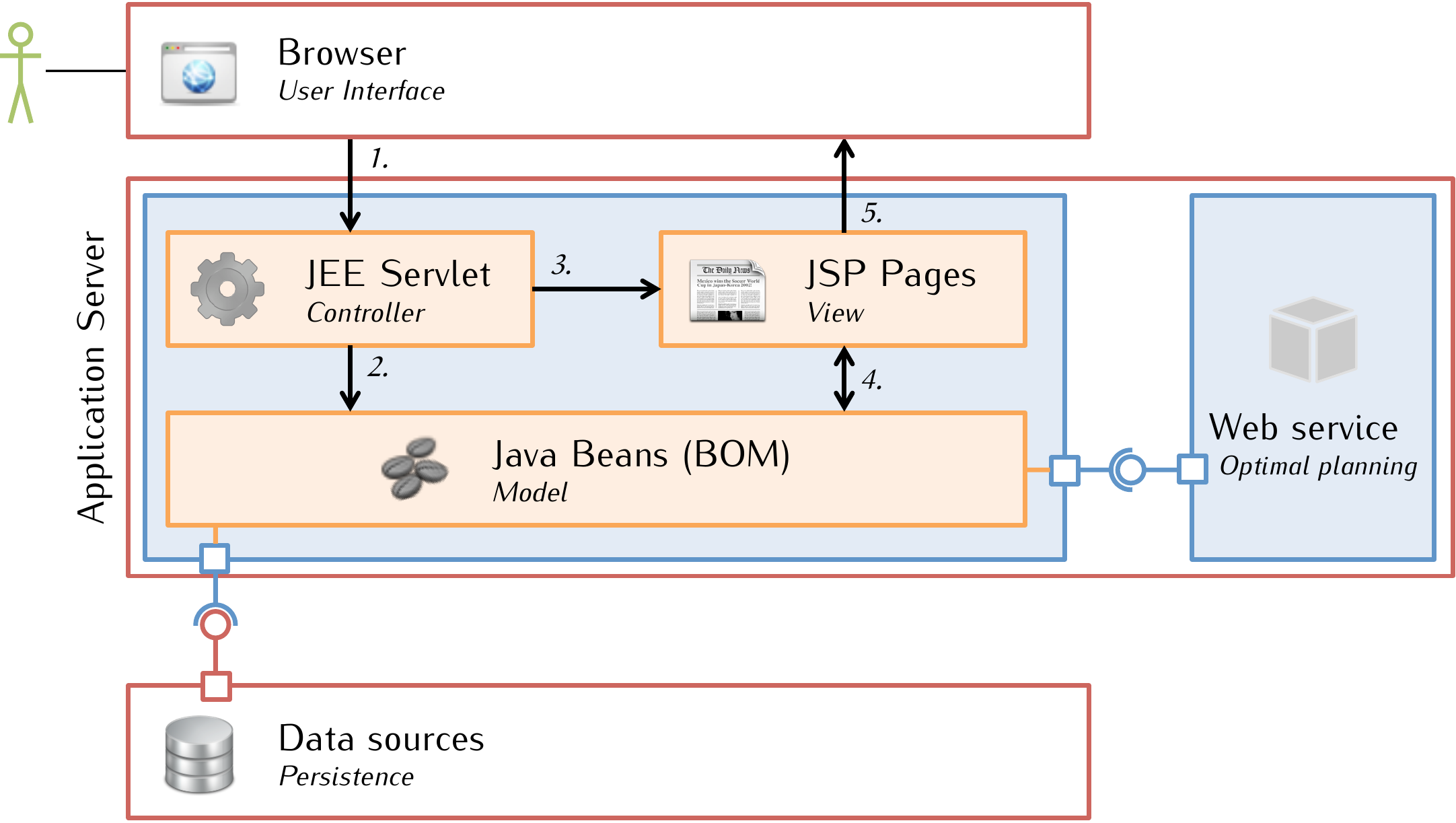}
                \caption{The Web application.}
				\label{fig:webapp}
        \end{subfigure}
        \caption{Software stack to deploy the \acs{CLP} program as a Web service and the typical \acs{MVC} pattern to exploit it as a Web application.}
		\label{fig:architecture}
\end{figure}

Usually, policy makers are not IT-experts. To ease the access to the DSS, we deployed the \ac{CLP} planner as a stateless Web service and access it by means of a stateful Web application.
The \ac{CLP} program is embedded inside a Java wrapper (\figurename~\ref{fig:service}) that encodes the requests in \ac{CLP} terms and decodes the results. This component provides a plethora of Java classes that represent the \ac{BOM} of this domain. Any query addressed to this component and all the returned  results are expressed in terms of these objects. We adopted the Apache CXF framework to support the Web Service imlpementation. The Web application that stands as a \ac{GUI} for the Web service is a standard Java servlet (\figurename~\ref{fig:webapp}) following the \ac{MVC} pattern and can be accessed at: \url{http://globalopt.epolicy-project.eu/Pareto/}.

After a welcome page that introduces the software, there are an input page, and a results page.
As input the user can provide minimum and maximum bounds for each energy source, constraints,
and objective functions for the Pareto optimization (together with a desired number of Pareto points).
Constraints and objectives can include linear combinations of cost, produced power, receptors, emissions, or indicators.

As a result, a set of graphs allow to inspect details on a specific plan (scenario), and/or to compare the computed plans.
Scenarios are divided into {\em boundary scenarios}, that are those that optimize one of the objective functions, and {\em intermediate scenarios}, that try to balance the various objectives.

\begin{figure}[tb]
		        \centering
		        \begin{subfigure}[b]{0.495\textwidth}
		                \includegraphics[width=\textwidth]{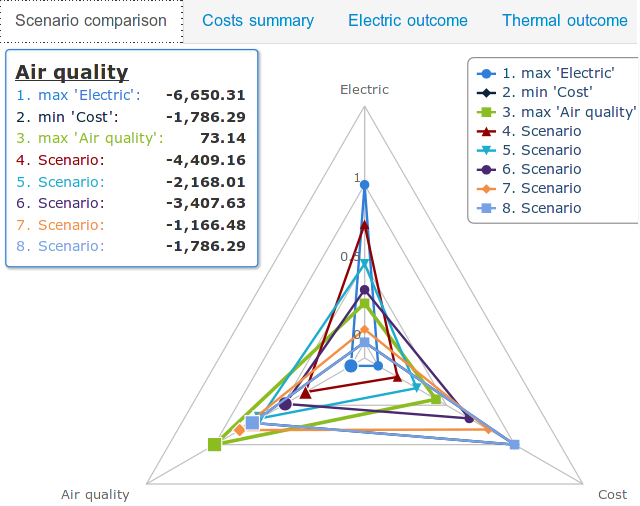}
		                \caption{Scenario comparison.}
						\label{fig:comparison}
		        \end{subfigure}%
		        \hfill %add desired spacing between images, e. g. ~, \quad, \qquad etc.
		        \begin{subfigure}[b]{0.495\textwidth}
		                \includegraphics[width=\textwidth]{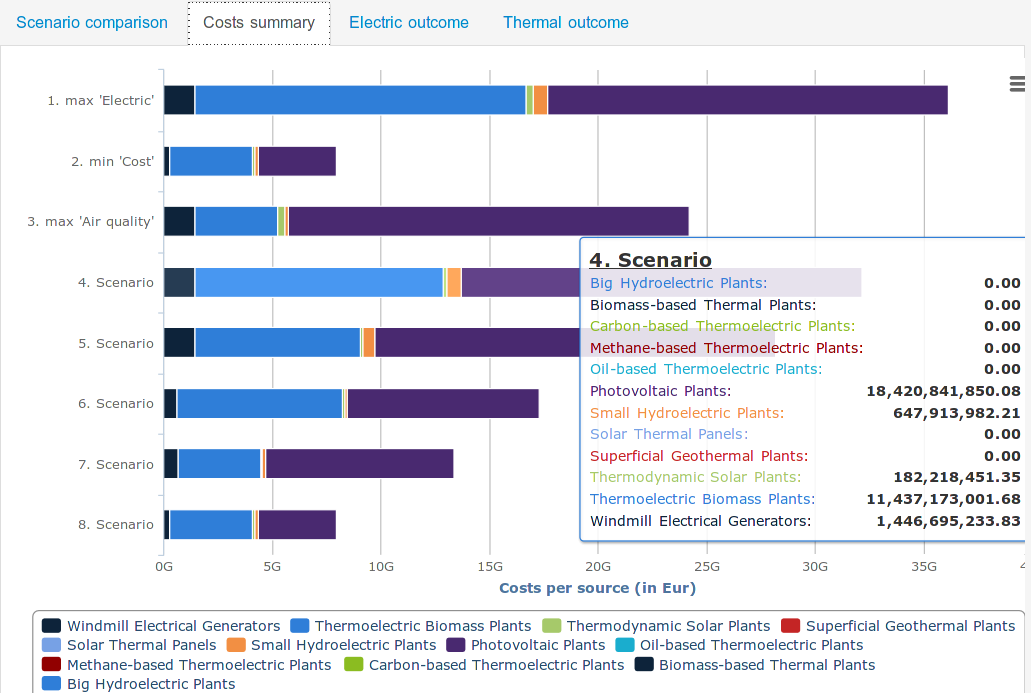}
		                \caption{Costs summary.}
						\label{fig:costs}
		        \end{subfigure}%
		        \caption{The views associated with the \emph{General overview} entry for \emph{Scenarios comparison}.}
				\label{fig:overview}
		\end{figure}

\paragraph{Scenarios comparison.}
Scenarios can be compared through a \emph{spiderweb chart} (\figurename~\ref{fig:comparison}) that has an axis for each objective function. Along each axis, the optimal values are far from the origin, and each scenario is represented by a polygon. Roughly speaking, a bigger polygon implies a better scenario (note that these solutions are Pareto optimal, so one polygon cannot be completely included into another polygon).

Scenarios can also be compared through \emph{stacked bar chart}, showing, for each scenario, the distribution of costs per energy source (\figurename~\ref{fig:costs}), or the amount of electric/thermal energy per source. Moreover, a further view provides scenarios comparison in terms of pollutants (\emph{Heavy metals}, \emph{Greenhouse gases,} and \emph{Other pollutants}), by means of basic column charts.

	\begin{figure}[tb]
	        \centering
	        \begin{subfigure}[b]{0.495\textwidth}
	                \includegraphics[width=\textwidth]{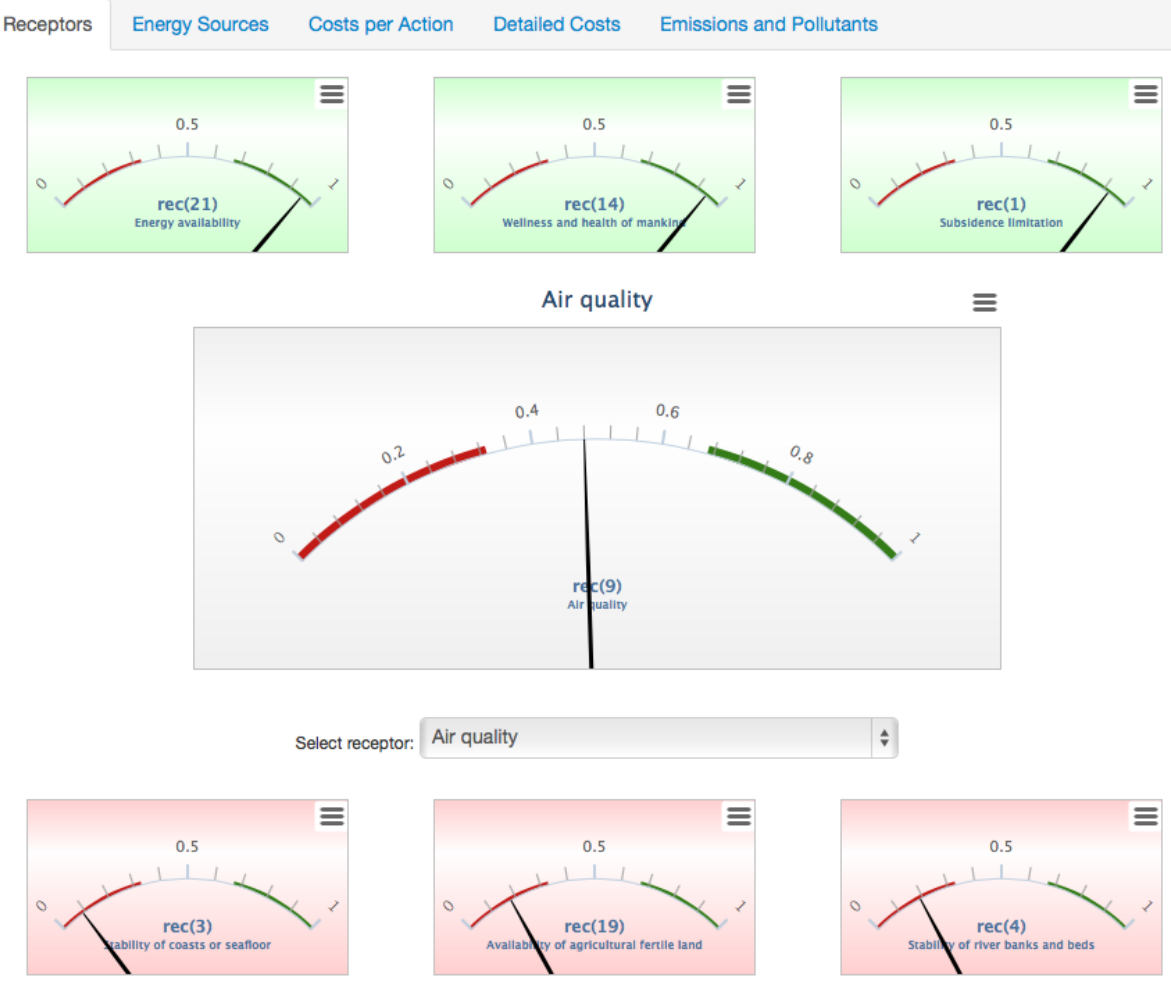}
	                \caption{Receptors.}
					\label{fig:receptors}
	        \end{subfigure}%
	        \hfill %add desired spacing between images, e. g. ~, \quad, \qquad etc.
	        \begin{subfigure}[b]{0.495\textwidth}
	                \includegraphics[width=\textwidth]{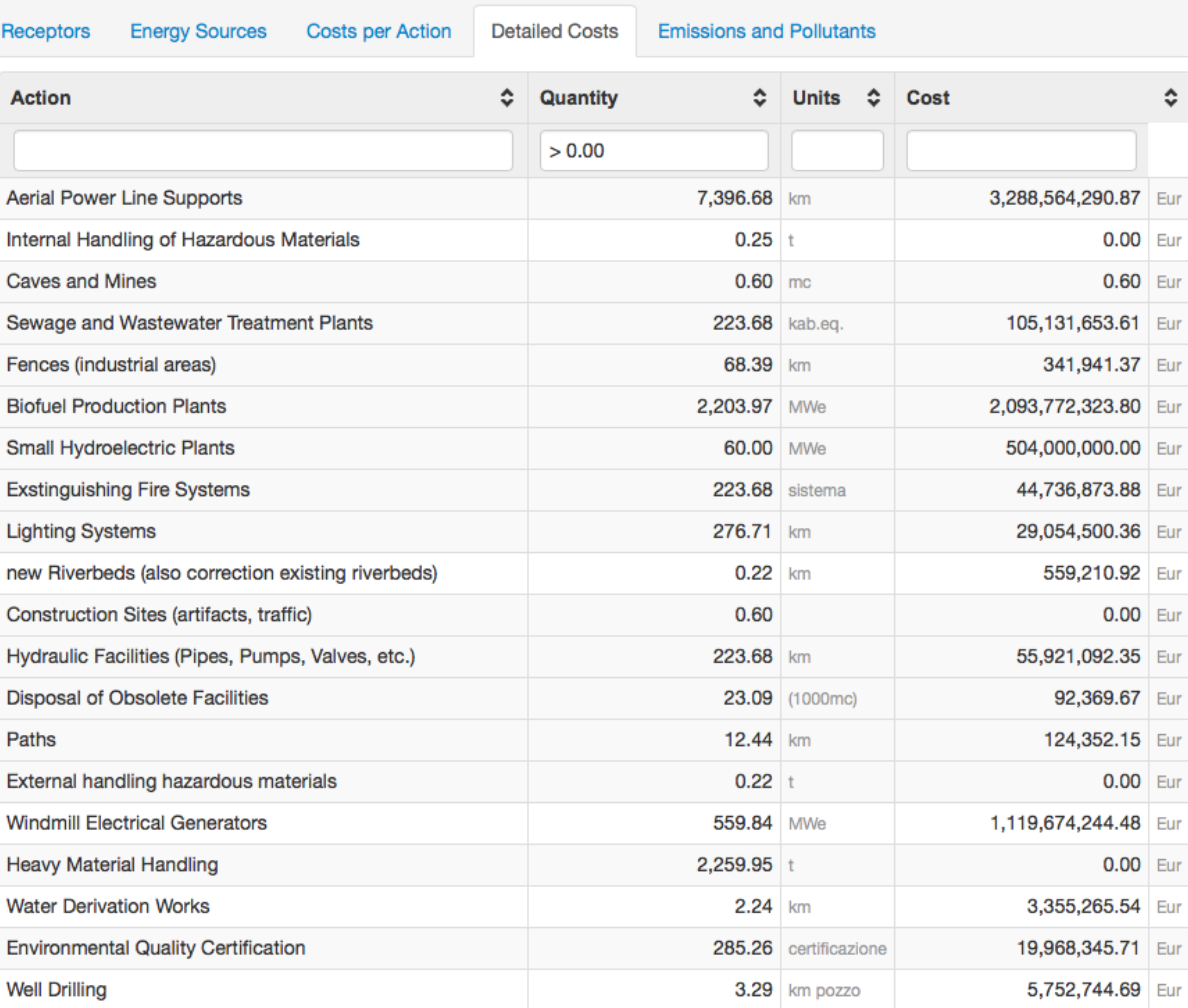}
	                \caption{A tabular view.}
					\label{fig:tabular}
	        \end{subfigure}
					\caption{Details views for each scenario.}
			\label{fig:scenario}
	\end{figure}

\paragraph{Scenarios Details.}
For each scenario, the following views are available:
		\begin{itemize}
			\item \emph{Receptors.} This composite view uses 7 \emph{VU\--meter charts} (\figurename~\ref{fig:receptors}). The top part shows the 3 receptors with the best normalised value, while the bottom one the 3 with the worst normalised value. The main chart allows the user to select any receptor and appraise its normalised value. This specific view ensures fast access to the best and worst receptors for the specific scenario.
			\item \emph{Other views. } There are four interactive \emph{tabular} views (\figurename~\ref{fig:tabular}) showing respectively, for the chosen scenario, the amount of produced energy per source, the total cost for each energy source to be spent in primary and secondary activities, the detailed costs for each activity, and the list of emissions.
		\end{itemize}

%%% Local Variables:
%%% TeX-master: "iclp14"
%%% End:

%% file: conclusions.tex
\section{Conclusions and Future Work}
\label{sec:conclusions}

We presented a decision support system with optimization based on \ac{CLP} for the regional planning, with particular emphasis on the environmental aspects.
The program was practically used to produce the energy plan 2011-2013 of the Emilia-Romagna region in Italy
\cite{PER2011-13}, and it is foreseen to use it also for the forthcoming plans.
The \ac{CLP} program is included into a standard web service, and it
has been equipped with an intuitive GUI.
The \ac{CLP} program will be the heart of the platform of the EU FP7 ePolicy project,
that will also include components like a social simulator, an opinion
miner, and a mechanism designer, all governed by the described
\ac{CLP} program. Preliminary work has been done on its integration
with the mechanism designer \cite{ECMS12}, and a social
simulator \cite{ECMS13}.

Future work will be on extending the model at a more detailed level,
e.g. taking into account decommissioning fixed costs. 

%%% Local Variables:
%%% TeX-master: "iclp14"
%%% End:

%% file: acronyms.tex
\begin{acronym}[ABCDEF]
	\acro{BOM}{Business Object Model}
%	\acro{CP}{Constraint Programming} %MG: Direi di non usare "CP", perche' siamo su una rivista di logica
	\acro{CLP}{Constraint Logic Programming}  
	\acro{DSS}{Decision Support System}
	\acro{CPS}{Constraint Programming System}
	\acro{GUI}{Graphical User Interface}
	\acro{HTML}{Hyper\--Textual Markup Language}
	\acro{JSP}{JavaServer Page}
	\acro{MVC}{Model\--View\--Controller}
	\acro{SaaS}{Software as a Service}
	\acro{SOA}{Service\--Oriented Architecture}
	\acro{UI}{User Interface}
	\acro{WSSEI}[WS\,SEI]{Web Service's Service Endpoint Interface}
	\acro{NOx}[$NO_X$]{Nitrogen Oxides}
	\acro{PAH}{Polycyclic Aromatic Hydrocarbon compounds}
	\acro{SOx}[$SO_X$]{Sulfur Oxides}
	\acro{NMVOC}{Non-Methane Volatile Organic Compounds}
	\acro{PCB}{Polychlorinated biphenyls}
	\acro{HCB}{Hexachlorobenzene}
\end{acronym}